\documentclass{sig-alternate}
\usepackage{url, array, multirow}
\newcolumntype{C}{>{\centering\arraybackslash}m{3.5em}}
\newcolumntype{K}{>{\centering\arraybackslash}m{12em}}

\begin{document}
\setlength{\paperheight}{11in}
\setlength{\paperwidth}{8.5in}
\pdfpagewidth=8.5in
\pdfpageheight=11in

\def\etal{\emph{et al.}~}
\newfont{\mycrnotice}{ptmr8t at 7pt}
\newfont{\myconfname}{ptmri8t at 7pt}
\let\crnotice\mycrnotice%
\let\confname\myconfname%

\permission{Permission to make digital or hard copies of all or part of this work for personal or classroom use is granted without fee provided that copies are not made or distributed for profit or commercial advantage and that copies bear this notice and the full citation on the first page. Copyrights for components of this work owned by others than ACM must be honored. Abstracting with credit is permitted. To copy otherwise, or republish, to post on servers or to redistribute to lists, requires prior specific permission and/or a fee. Request permissions from Permissions@acm.org.}
\conferenceinfo{MM'15,}{October 26--30, 2015, Brisbane, Australia.} 
\copyrightetc{\copyright~2015 ACM. ISBN \the\acmcopyr}
\crdata{978-1-4503-3459-4/15/10\ ...\$15.00.\\
DOI: http://dx.doi.org/10.1145/2733373.2806316}

\clubpenalty=10000 
\widowpenalty = 10000

\title{A Deep Siamese Network\\for Scene Detection in Broadcast Videos}

\numberofauthors{3} 

\author{
%
%
Lorenzo Baraldi, Costantino Grana and Rita Cucchiara\\
       \affaddr{Dipartimento di Ingegneria ``Enzo Ferrari'', Universit\`a degli Studi di Modena e Reggio Emilia}\\
       \affaddr{Via P. Vivarelli, 10, Modena MO 41125, Italy}\\
       \email{name.surname@unimore.it}
}
\maketitle
\begin{abstract}
We present a model that automatically divides broadcast videos into coherent scenes by learning a distance measure between shots. Experiments are performed to demonstrate the effectiveness of our approach by comparing our algorithm against recent proposals for automatic scene segmentation. We also propose an improved performance measure that aims to reduce the gap between numerical evaluation and expected results, and propose and release a new benchmark dataset.
\end{abstract}




\section{Introduction} 
Scene detection is the task to automatically segment an input video into meaningful and story-telling parts, without any help from the producer, using perceptual cues and multimedia features extracted from data~\cite{vendrig02}. Therefore, it can be an effective tool to enhance video accessing and browsing; moreover, since each of the resulting parts could be automatically tagged, this kind of decomposition can enable a finer-grained search inside videos.

\begin{figure}[tbh]
\centering
\includegraphics[width=0.94\linewidth]{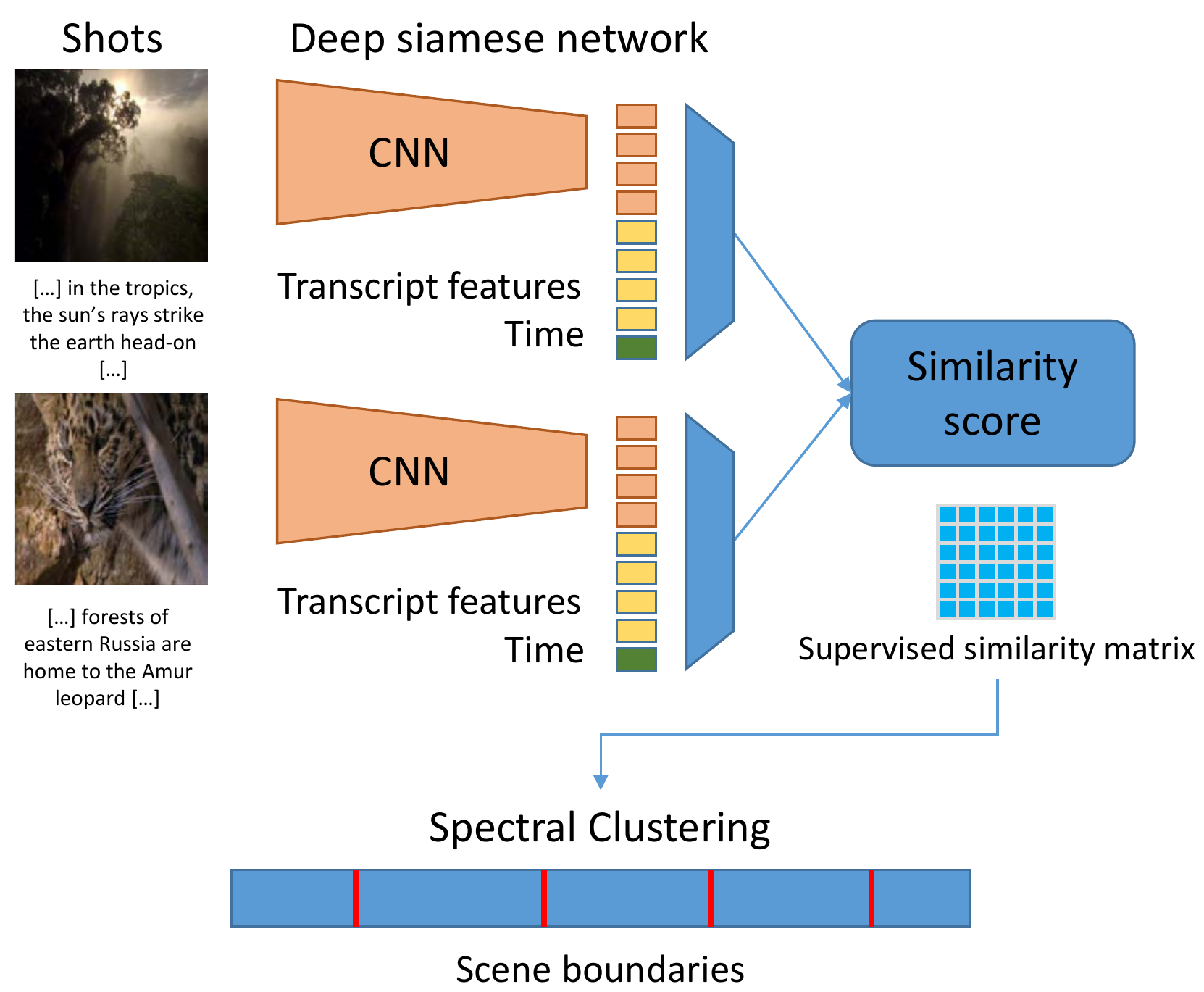}
\caption{Our approach decomposes a video into coherent parts. A multimodal deep neural network learns a similarity score for each pair of shots, using visual and textual features. The resulting similarity matrix is used to cluster adjacent shots together.}
\label{fig:approach}
\end{figure}

We address the problem of automatic scene detection in broadcast video. Differently from news videos, which present a well established structure, generic broadcast videos have different editing standards based on the specific style the director desires. This cue has been exploited in existing works: Liu \etal \cite{liu13}, for example, propose a probabilistic framework that imitates the authoring process and detects scenes by learning a scene model. 
Another solution is instead to disregard the structure, focusing only on similarities: in~\cite{chasanis09}, shots are firstly clustered into symbolic groups, then, scene boundaries are detected by comparing successive non-overlapping windows of shot labels using a sequence alignment technique that considers the visual similarity of shot clusters and the frequency of sequential labels in the video.

An alternative view of the scenes structure is to arrange shots in a graph representation and then cluster them by partitioning the graph. Sidiropoulos \etal \cite{sidiropoulos11}, for example, exploited the Shot Transition Graph method, where each node represents a shot and edges between shots are weighted by shot similarity. All these techniques rely on well established features such as histograms, bag-of-words representations, or MPEG-7 descriptors, extracted from one or multiple key frames; audio or transcript features have also been employed.

The task of scene detection requires good representations of the video content, specifically images and transcript. Recently, Convolutional Neural Networks have shown their powerful abilities on image representation~\cite{krizhevsky12}. In this paper we go beyond traditional hand-crafted features and apply the deep learning paradigm to scene detection, exploiting both visual and textual features from the transcript, that, when not directly available, can be obtained with automatic speech recognition. Deeply learned features are then used together with a clustering algorithm to segment the video. To the best of our knowledge, this is the first attempt to use deep learning in this task.

As it usually happens in emerging topics, most works on scene detection conducted experiments on personal test sets, which are not publicly available, thus making it hard for others to reproduce or to compare the presented results. Evaluation measures were also sometimes not appropriate and did not reflect the quality perceived by the user. For these reasons, we propose a new dataset for scene detection, and we also try to tackle the problem of evaluation.

The main contributions of our work are: 
\begin{itemize}
\item We propose a deep learning framework for segmenting videos into coherent scenes, which takes both visual and textual features as input and merges them to create similarity scores;
\item A new benchmark dataset is proposed, and annotations are publicly released. To our knowledge, this is the biggest dataset for scene detection available.
\item Lastly, we address the limitations of existing measures for scene detection evaluation, and propose an improved measure which solves frequently observed cases in which the numeric interpretation would be different from the expected results.
\end{itemize}

\section{Deeply-Learned Scene Detection}
Given the nature of a broadcast video -- i.e. a sequence of shots, and given that a shot usually has a uniform content, scene detection can also be viewed as the problem of grouping adjacent shots together, with the objective of maximizing the semantic coherence of the resulting segments. This implies a dimensionality reduction and poses scene detection as a clustering problem: our model, indeed, detects scenes by applying the spectral clustering algorithm to shots. Moreover, for the detected segmentation to be useful to the final user, we want it to be as close as possible to the desired output. For this reason, instead of applying pre-defined descriptors to build the similarity matrix the clustering algorithm needs, we couple the unsupervised clustering with a supervised deep neural network which learns similarities between shots.

The architecture of our neural network resembles that of a Siamese network~\cite{hadsell06} (see Fig. \ref{fig:approach}): it consists of two branches that share exactly the same architecture and the same weights. Each branch takes as input two distinct shots, and then applies a series of convolutional, ReLU and max-pooling layers. Branch outputs are then concatenated and given to a top network. Branch of the Siamese network can be seen as descriptor computation modules and the top network as a similarity function. At test time, similarity scores computed by the neural network are composed together to build a similarity matrix, which is then given to the spectral clustering algorithm to obtain the final scene boundaries.

\subsection{Visual features}
The first part of each branch consists of a Convolutional Neural Network (CNN) which takes the middle frame of a given shot, cropped and resized to fit the network input size. Using a single frame allows a complexity reduction while still allowing a good description.

We want to give the CNN the ability to recognize not only objects, animals and people, but indoor and outdoor places too, since a scene often takes place in a single location, and to be specific for the scene segmentation task. Therefore, we pre-train our CNN on 1.2 million images of the ImageNet dataset (ILSVRC 2012) \cite{krizhevsky12}, plus 2.5 million images from the Places dataset~\cite{zhou14}. Finally, the CNN is fine-tuned using training shots.

The architecture of the CNN is the same as the one used in the Caffe reference network~\cite{jia14}, and the visual representation of the shot is computed as follows:
\begin{equation}
r_{vis} = \sigma(\mathbf{w}_{vis}(CNN_{vis}(I)) + b_{vis}
\end{equation}
where $\sigma(\cdot)$ is the ReLU activation function. The image CNN returns the 4096-dimensional activations of the fully connected layer immediately before the last ReLU layer. The matrix $\mathbf{w}_{vis}$ has dimension $d \cdot 4096$, and each image is thus represented as one $d_{vis}$-dimensional vector $r_{vis}$.

\subsection{Textual features}
Beside the visual content of a shot, we want to take into account the content of the transcript, while still keeping a shot-based representation. Given that a shot can contain a variable number of words, and a feed-forward neural network requires fixed size inputs, we exploit a variant of the bag-of-words approach that takes feature vectors obtained with Skip-gram models~\cite{mikolov13}: words in the transcript are represented using their Word2vec descriptors, and then clustered using \textit{k}-means and the cosine distance between words. 

Since shots can be very short, describing a brief shot using only the words it contains would result in a non-consistent descriptor. Therefore, for each shot we define a context window centered on the shot center frame, and with size $\max(w_s, W)$, where $w_s$ is the shot duration and $W$ is the minimum context window size. Words from each window are then represented with their bag-of-words vector, resulting in a $d_{words}$-dimensional vector $r_{words}$ for each shot.

\subsection{Similarity scores and clustering}
\begin{figure}
\centering
\includegraphics[width=0.54\linewidth]{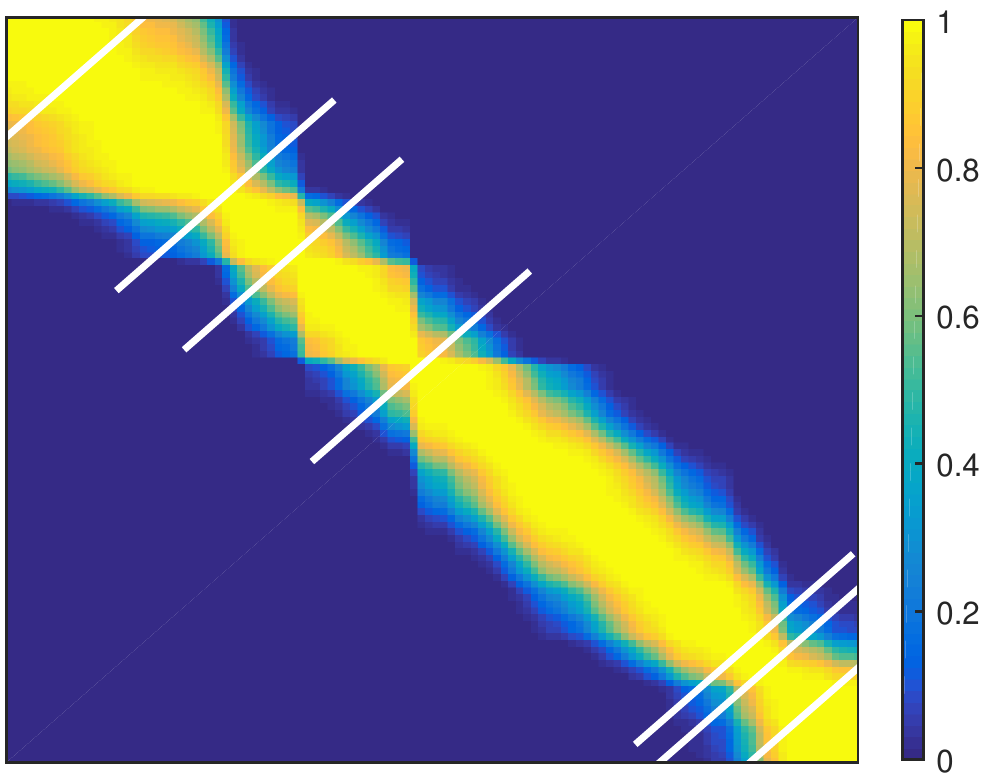}
\caption{Detail of matrix $W$ for the \textit{From Pole to Pole} episode. Ground truth scene boundaries (white lines) correspond to low similarity areas.}
\label{fig:W}
\end{figure}
In the last part of each branch, the visual representation $r_{vis}$ is merged with the textual representation $r_{words}$; in addition, the resulting vector is concatenated with the index of the center frame of the shot, so that the network is aware of the temporal distance between two shots. A fully connected layer takes the joint representation and learns how to weight the components to get the final similarity scores. 

The network is trained using a contrastive loss term and squared $l_2$-norm regularization, that leads to the following learning objective function:
\begin{equation}
L(\mathbf{w}) = \frac{\lambda}{2} \|\mathbf{w}\|^2_2 + \frac{1}{2N} \sum_{(i,j) \in \mathcal{D}} y_{ij} d_{ij}^2  + (1-y_{ij}) \max(1-d_{ij}^2, 0)
\end{equation}
where $\mathbf{w}$ are the weights of the neural network, $\mathcal{D}$ is the set of training shot pairs, $d_{ij}^2$ is the squared $l_2$-distance for shots $i$ and $j$ (computed between the two final layers of the Siamese network), and $y_{ij} \in \{ 0, 1\}$ is the corresponding label (with $0$ and $1$ denoting a non-matching and a matching pair, respectively).

Finally, distances $d_{ij}$ are turned into similarity scores by applying a Gaussian kernel, where bandwith $\sigma$ is computed using a kernel density estimator. Similarity matrix $W$ (see Fig.~\ref{fig:W} for an example) is then used together with spectral clustering to group adjacent shots: final scene boundaries are placed between shots belonging to different clusters.
 
\section{Experimental evaluation}
\textbf{Performance measures}~~
The problem of measuring scene detection performance is significantly different from that of measuring shot detection performance. Indeed, classical boundary detection scores, such as Precision and Recall, fail to convey the true perception of an error, which is different for an off-by-one shot or for a completely missed scene boundary.

In~\cite{vendrig02} the Coverage and Overflow measures were proposed to overcome this limitation. Coverage $\mathcal{C}$ measures the quantity of shots belonging to the same scene correctly grouped together, while Overflow $\mathcal{O}$ evaluates to what extent shots not belonging to the same scene are erroneously grouped together. Formally, given the set of automatically detected scenes $\mathbf{s} = \left[ \mathbf{s}_1, \mathbf{s}_2, ..., \mathbf{s}_m \right]$, and the ground truth $\tilde{\mathbf{s}} = \left[ \tilde{\mathbf{s}}_1, \tilde{\mathbf{s}}_2, ..., \tilde{\mathbf{s}}_n \right]$, where each element of $\mathbf{s}$ and $\tilde{\mathbf{s}}$ is a set of shot indexes, the coverage $\mathcal{C}_t$ of scene $\tilde{\mathbf{s}}_t$ is defined as:
\begin{equation}
\mathcal{C}_t = \frac{\max_{i=1...,m} \#(\mathbf{s}_i \cap \tilde{\mathbf{s}}_t)}{\#(\tilde{\mathbf{s}}_t)}
\end{equation}
where $\#(\mathbf{s}_i)$ is the number of shots in scene $\mathbf{s}_i$. The overflow of a scene $\tilde{\mathbf{s}}_t$, $\mathcal{O}_t$, is the amount of overlap of every $\mathbf{s}_i$ corresponding to $\tilde{\mathbf{s}}_t$ with the two surrounding scenes:
\begin{equation}
\mathcal{O}_t = \frac{\sum_{i=1}^{m} \#(\mathbf{s}_i \setminus \tilde{\mathbf{s}}_t)\cdot \min(1,\#(\mathbf{s}_i \cap \tilde{\mathbf{s}}_t))}{\#(\mathbf{\tilde{s}}_{t-1})+\#(\tilde{\mathbf{s}}_{t+1})}
\end{equation}
The per-ground-truth-scene measures are aggregated for the entire video by averaging, weighting them by the number of shots in each scene. Finally, an F-Score measure, $F_{co}$, can be defined to combine Coverage and Overflow in a single measure, by taking the harmonic mean of $\mathcal{C}$ and $1-\mathcal{O}$.

These measures, while going in the right direction, have a number of drawbacks, which may affect the evaluation. As also noted in~\cite{sidiropoulos12}, $F_{co}$ is not symmetric, leading to unusual phenomena in which an early or late positioning of the scene boundary, of the same amount of shots, may lead to strongly different results. Moreover, the relation of $\mathcal{O}$ with the previous and next scenes creates unreasonable dependencies between an error and the length of a scene observed many shots before it. Finally, $\mathcal{C}$ only depends on the maximum overlapping scene, and does not penalize the other overlapping scenes in any way: any over-segmentation in the other overlapping scenes does not change the measure value.

We propose to use a symmetric measure based on intersection over union to assess the quality of detected scenes. For each ground-truth scene, we take the maximum intersection-over-union with the detected scenes, averaging them on the whole video. Then the same is done for detected scenes against ground-truth scene, and the two quantities are again averaged. An important note is that both intersection and union are measured in terms of frame lengths for the shots, thus weighting the shots with their relative significance. The final measure is thus given by:
\begin{equation}
M_{iou} = \frac{1}{2} \left(
\frac{1}{n} \sum_{i=1}^n \max_{j \in \mathbb{N}_m} \frac{\tilde{\mathbf{s}_i} \cap \mathbf{s}_j}{\tilde{\mathbf{s}_i} \cup \tilde{\mathbf{s}}_j} + 
\frac{1}{m} \sum_{j=1}^m \max_{i \in \mathbb{N}_n} \frac{\tilde{\mathbf{s}_i} \cap \mathbf{s}_j}{\tilde{\mathbf{s}_i} \cup \tilde{\mathbf{s}}_j}
\right)
\end{equation}

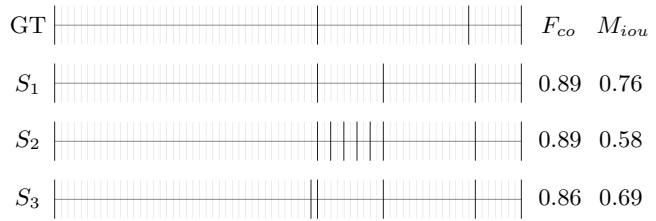
\begin{figure}
	\centering
	\begin{picture}(260,90)
		\put(15,0){\includegraphics[width=180px]{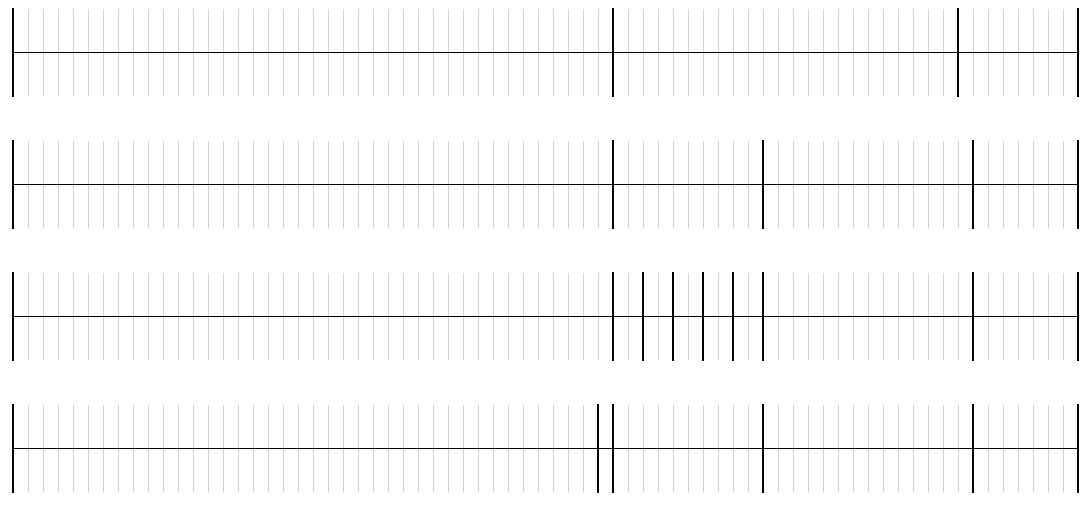}}
		\put(0,72){GT}  \put(201,72){$F_{co}$}	\put(222,72){$M_{iou}$}
		\put(2,50){$S_1$} \put(200,50){$0.89$}		\put(223,50){$0.76$}
		\put(2,29){$S_2$}	\put(200,29){$0.89$}		\put(223,29){$0.58$}
		\put(2,7){$S_3$} 	\put(200,7){$0.86$} 		\put(223,7){$0.69$}
	\end{picture}
	\caption{Values of $F_{co}$ and $M_{iou}$ for sample scene detections. Gray and black ticks represent shot and scene boundaries respectively.}
	\label{fig:esempio}
\end{figure}

Figure~\ref{fig:esempio} shows the behavior of the two measures in three synthetic cases. As it can be seen, $M_{iou}$, unlike $F_{co}$, penalizes over-segmentations and does not create dependencies between an error and previous scenes. 

\medskip
\textbf{Experiments setting}~~
We evaluate our method on 11 episodes from the BBC educational TV series \textit{Planet Earth}\footnote{\url{http://www.bbc.co.uk/programmes/b006mywy}}. Each episode is approximately 50 minutes long, and the whole dataset contains around 4900 shots and 670 scenes. Shots and scenes of the entire dataset have been manually annotated by a set of human experts: annotations, as well as the Caffe~\cite{jia14} models of our network, are available at \url{http://imagelab.ing.unimore.it}.

To train our model, we employ Stochastic gradient descent with momentum $0.9$ and weight decay $\lambda = 0.0005$. The learning rate is set to $0.001$ for CNN neurons, and to $0.004$ for the others. Parameters $d_{vis}$, $d_{words}$ and $W$ are set to 1183, 200 and 20 seconds, respectively, and the last fully connected layer of each branch is composed by 200 neurons. Training is done in mini-batches of size $128$, and we shuffle and augment training data by flipping both shots horizontally and vertically. We also subtract from each frame the average frame computed over the training set. 

The training set for a given video corresponds to all possible pairs of shots, most of them not belonging to the same scene, and it is therefore heavily unbalanced. To avoid the risk of having batches with only negative examples, and to balance the training phase, we artificially build batches using the same amount of positive and negative examples.

\begin{table*}[t]
\centering
\caption{Results on BBC \textit{Planet Earth} series, using Intersection-over-union and Coverage-Overflow measures}
\begin{tabular}{|K|C|C|C|C||C|C|C|C|} 
\hline
\multirow{2}{*}{\textbf{Episode}} & \multicolumn{4}{c||}{$M_{iou}$} & \multicolumn{4}{c|}{$F_{co}$} \\ 
                        \cline{2-9}
			&STG~\cite{sidiropoulos11} & Color + NW\cite{chasanis09} & Color + SC & Our method & STG~\cite{sidiropoulos11} & Color + NW\cite{chasanis09} & Color + SC & Our method \\  \hline
From Pole to Pole       & 0.42          & 0.35          & 0.43          & \textbf{0.50} & 0.47          & 0.39          & 0.48          & \textbf{0.56} \\  \hline 
Mountains               & 0.40          & 0.31          & 0.44          & \textbf{0.53} & 0.55          & 0.51          & 0.54          & \textbf{0.63} \\  \hline 
Fresh Water             & 0.39          & 0.34          & 0.48          & \textbf{0.52} & 0.56          & 0.53          & 0.50          & \textbf{0.66} \\  \hline 
Caves                   & 0.37          & 0.33          & 0.44          & \textbf{0.55} & 0.46          & 0.39          & 0.42          & \textbf{0.61} \\  \hline 
Deserts                 & 0.36          & 0.33          & \textbf{0.46} & 0.36          & 0.42          & \textbf{0.56} & 0.54          & 0.55          \\  \hline 
Ice Worlds              & 0.39          & 0.37          & 0.44          & \textbf{0.51} & 0.44          & 0.45          & 0.50          & \textbf{0.64} \\  \hline 
Great Plains            & 0.46          & 0.37          & \textbf{0.48} & 0.47          & \textbf{0.65} & 0.53          & 0.51          & 0.59          \\  \hline 
Jungles                 & 0.45          & 0.38          & \textbf{0.53} & 0.51          & 0.63          & 0.45          & 0.60          & \textbf{0.64} \\  \hline 
Shallow Seas            & 0.46          & 0.32          & 0.47          & \textbf{0.51} & 0.61          & 0.36          & 0.55          & \textbf{0.64} \\  \hline 
Seasonal Forests        & 0.42          & 0.20          & \textbf{0.43} & 0.38          & 0.58          & 0.19          & 0.52          & \textbf{0.64} \\  \hline 
Ocean Deep              & 0.34          & 0.36          & \textbf{0.48} & \textbf{0.48} & 0.52          & 0.54          & 0.57          & \textbf{0.64} \\  \hline 
\hline
\textbf{Average}        & 0.41          & 0.33          & 0.46          & \textbf{0.48} & 0.54          & 0.45          & 0.52          & \textbf{0.62} \\  \hline 
\hline
\textbf{Average with GT shots} &  --    & 0.39          & 0.47          & \textbf{0.51} &  --           & 0.55          & 0.50          & \textbf{0.62} \\  \hline
\end{tabular}
\label{tab:results}
\end{table*}

\medskip
\textbf{Evaluation}~~
We compare our model against two recent algorithms for scene detection: \cite{sidiropoulos11}, which uses a variety of visual and audio features merged in a Shot Transition Graph (STG), and \cite{chasanis09}, that combines low level color features with the Needleman-Wunsh (NW) algorithm. We further include a baseline approach, which clusters shots using spectral clustering (SC) and three-dimensional color histograms and time as features. We use the executable of~\cite{sidiropoulos11} provided by the authors and reimplemented the method in~\cite{chasanis09}. Training of the deep neural network was performed in a leave-one-out setup (ten videos for training and one for testing), and parameters of all methods were selected to maximize the performance on the training set. Since the performance of shot detection can condition the performance of scene detection, all experiments were carried out using the shot detector in~\cite{apostolidis14}, which is the same exploited by the executable of~\cite{sidiropoulos11}.

Table~\ref{tab:results} shows experimental results on the \textit{BBC Planet Earth} series, using both measures. Bottom line reports scene detection results when using ground truth shot boundaries instead of those obtained with~\cite{apostolidis14}. Reported performances clearly show that color features, when used in combination with spectral clustering, can achieve good results according to both measures. The color histograms baseline, indeed, is superior or equivalent to the STG approach in~\cite{sidiropoulos11} and to that of~\cite{chasanis09}. Our full model, which exploits both visual and textual learned features, shows consistent improvement over the baseline and over both the approaches it has been compared to.

\section{Conclusions}
We introduced a deep learning architecture that merges visual and textual data to partition broadcast videos into coherent parts. We showed that this model provides state of the art performance when compared to recent proposals for scene detection, both with classical performance measures, and with an improved proposal.

\section*{Acknowledgments}
This work was carried out within the project ``Citt\`a educante'' (CTN01\_00034\_393801) of the National Technological Cluster on Smart Communities cofunded by the Italian Ministry of Education, University and Research - MIUR.

\bibliographystyle{abbrv}
\bibliography{sigproc}

\end{document}